\documentclass[11pt,a4paper]{article}
\usepackage[hyperref]{acl2020}
\usepackage{times}
\usepackage{latexsym}

\usepackage{microtype}

\aclfinalcopy 

\usepackage{url}

\usepackage{bbm}
\usepackage{microtype} 
\usepackage{graphicx}
\usepackage{subfigure}
\usepackage{booktabs} 
\usepackage{amssymb}
\usepackage{amsthm}
\usepackage{mathtools}
\usepackage{multirow}
\usepackage{soul}

\usepackage{enumitem}
\setlist{nolistsep}

\title{Detecting Domain Polarity-Changes of Words in a Sentiment Lexicon}

\author{Shuai Wang$^{\dag}$, Guangyi Lv$^{\ddag}$, Sahisnu Mazumder$^{\dag}$,  Bing Liu$^{\dag}$\\
$^{\dag}$Department of Computer Science,  University of Illinois at Chicago, USA\\
$^{\ddag}$University of Science and Technology of China, Hefei, China \\
  \texttt{shuaiwanghk@gmail.com, gylv@mail.ustc.edu.cn} \\
    \texttt{sahisnumazumder@gmail.com, liub@uic.edu} 
  }

\date{}

\begin{document}
\maketitle
\begin{abstract}
Sentiment lexicons are instrumental for sentiment analysis. One can use a set of sentiment words provided in a sentiment lexicon and a lexicon-based classifier to perform sentiment classification. One major issue with this approach is that many sentiment words are domain dependent. That is, they may be positive in some domains but negative in some others. We refer to this problem as \textit{domain polarity-changes of words}. Detecting such words and correcting their sentiment for an application domain is very important. In this paper, we propose a graph-based technique to tackle this problem. Experimental results show its effectiveness on multiple real-world datasets.
\end{abstract}

\section{Introduction}
Sentiment words, also called opinion/polar words, are words that convey positive or negative sentiments~\cite{Pa:08}. Such sentiment-bearing words are usually pre-compiled as word lists in a sentiment lexicon, which is instrumental as well as an important linguistic resource to sentiment analysis~\cite{liu2012sentiment}. So far, numerous studies about how to construct lexicons have been reported, which will be discussed in Section~\ref{sec:related_work}. 

Despite the fact that there is extensive research on lexicon construction, limited work has been done to solve the problem of identifying and handling sentiment words in a lexicon that have domain-dependent polarities. In real-life applications, there are almost always some sentiment words that express sentiments different from their default polarities provided in a general-purpose sentiment lexicon. For example, in the sentiment lexicon compiled by \citet{Hu:04}, the word ``crush'' is associated with a negative sentiment, but it actually shows a positive opinion in domain \textit{blender} because ``crush'' indicates that a blender works well, e.g., in the sentence ``it does crush the ice!". We call this problem \textit{domain polarity-change} of a word in a sentiment lexicon. 

The polarity change of words plays a crucial role in sentiment classification. As we will see in the experiment section, without identifying and correcting such domain dependent sentiment words, sentiment classification performance could be much inferior. 
Although some researchers have studied the domain-specific sentiment problem with sentiment lexicons, their focuses are quite different and their approaches are not suitable for our task.
 We will discuss them further in the following sections.

It is important to note that our work mainly aims to help \textit{lexicon-based sentiment classification approaches} ~\cite{taboada2011lexicon}. It does not directly help machine-learning (ML) or supervised learning approaches~\cite{li2018hierarchical} because the domain-dependent polarities of words are already reflected in the manually labeled training data. Notice that for those ML approaches, the manual annotation for each application domain is required, which is a time-consuming and labor-intensive task, and is thus hard to scale up. In many real-world scenarios, lexicon-based approaches are useful and could be a better alternative~\cite{liu2012sentiment}.  

However, to effectively apply a sentiment lexicon for an application domain, the domain-polarity change problem discussed above needs to be addressed.
To this end, we propose a new graph-based approach named Domain-specific Sentiment Graph (DSG). It works with three main steps: (domain) sentiment words collection, (domain) sentiment correlation extraction, and graph construction and inference, which will be detailed in Section~\ref{sec:technique}.
Our experimental results show its effectiveness in detecting domain polarity changes of words on multiple real-world datasets. We will also see with the detection of the sentiment polarities of those words, a huge performance gain can be achieved in sentiment classification.

\section{Related Work}
\label{sec:related_work}
This work concerns about domain polarity changes of words in lexicons. So we first discuss the works related to sentiment lexicons, and then domain sentiment, and finally domain sentiment with lexicons. 

Extensive studies have been done for sentiment lexicons and the majority of them focus on lexicon construction. These approaches can be generally categorized as dictionary-based and corpus-based. 

Dictionary-based approaches first used some sentiment seed words to bootstrap based on the synonym and antonym structure of a dictionary \cite{Hu:04,Va:04}. Later on, more sophisticated methods were proposed 
\cite{Ki:04,Es:05,Ta:07,Bl:08,Ra:09,Mo:09,Ha:10,Dr:10,Xu:10,Pe:11,Ga:12,san2014simple}. 
Corpus-based approaches build lexicons by discovering sentiment words in a large corpus. The first idea is to exploit some coordinating conjunctions
~\cite{Ha:97,Ha:10}.
\citet{Kana:06} extended this approach by introducing inter-sentential sentiment consistency. Other related work includes~\cite{Ka:04,Ka:06,wang2017sentiment}. The second idea is to use syntactic relations between opinion and aspect words
~\cite{Zh:06,Wa:08,Qi:11,Vo:13}. The third idea is to use word co-occurrences for lexicon induction~\cite{Tu:03,igo2009corpus,Ve:10,yang2014topic,rothe2016ultradense}. 

However, our work is very different as we focus on detecting domain dependent sentiment words in a given general-purpose sentiment lexicon.

Also related is the existing research about domain and context dependent sentiment. First, despite the fact that several researchers have studied context dependent sentiment words, which are based on specific sentences and topic/aspect context~\cite{Wi:05,Di:08,Ch:08,Wu:10,Ji:10,Lu:11,Zh:12,Ke:12,teng2016context,wang2018target,li2018transformation}, our work is based on domains. 
Second, while the studies on transfer learning or domain adaptation for sentiment analysis deals with domain information~\cite{bhatt2015iterative,yu2016learning,li2018hierarchical}, our work does not lie in this direction. We do not have any source domain and our goal is not to transfer domain knowledge to another domain. Third, most importantly, the above works are either irrelevant to lexicons or not for detecting the sentiment discrepancy problem found in lexicons towards a particular domain.

Our work is most related to the following studies that involve both the sentiment lexicons and domain sentiment problem. 
\citet{Ch:09} adapted the word-level polarities of a general-purpose sentiment lexicon to a particular domain by utilizing the expression-level polarities in that domain. However, their work targeted at reasoning the sentiment polarities of multi-word expressions. It does not detect or revise the sentiment polarities of individual words in the lexicon for a particular domain, and hence, cannot solve our problem. \citet{Du:10} studied the problem of adapting the sentiment lexicon from one domain to another domain. It further assumes that the source domain has a set of sentiment-labeled reviews. Their technique is therefore more about transfer learning and their learning settings differ from ours intrinsically. 
Perhaps, the most related work is~\cite{hamilton2016inducing}, which generates domain-specific lexicons using some seed lexicon words, word embeddings, 
and the random walk algorithm. 
However, their model is primarily for lexicon construction, with domain-specific information involved/guided. 
It does not aim to detect/change the sentiment polarity from a given lexicon. It is thus not directly applicable to our task. To make it workable, we design a two-step approach, which will be detailed in the experiment section (Section~\ref{sec:experiments}).

\section{Problem Definition}
We first give the formal problem definition of detecting domain polarity-changes of words in a lexicon. \textbf{Definition: } given a general-purpose sentiment lexicon $L$ (the lexicon containing sentiment words and their default sentiment polarities) and an application domain review corpus $D$, to identify a subset of words in $L$ that have different sentiment polarities in that domain (different from their default polarities), which we call \textit{polarity-changed sentiment words} and denote them as $C$ $( C \subseteq L) $.

In the rest of this paper, we call the words in a given lexicon \textit{lexical words} for short. The term \textit{detection} will generally stand for the detection of domain polarity-changes of words. 

\section{Proposed Solution}
\label{sec:technique}
To tackle the above problem, we propose a graph-based learning approach named Domain-specific Sentiment Graph (DSG). It works with three major steps: (1) (domain) sentiment words collection, (2) (domain) sentiment correlation extraction, and (3) graph construction and inference.

Specifically, it first collects a set of mentioned sentiment words $S$ $(L \subseteq S)$ in the domain corpus $D$. It then mines multiple types of relationships among sentiment words in $S$, which are denoted as a relationship set $R$. The relationships are identified based on different types of linguistic connectivity. Next, it builds a probabilistic graph 
with each node representing a sentiment word in $S$ and each edge representing a relation (from $R$) between two words. An inference method is then applied to re-estimate the domain-specific polarities (or beliefs) of sentiment words. With the re-estimated beliefs obtained in the application domain, those sentiment words with changed polarities can be detected, based on the sentiment shift of a lexical word between its induced (in-domain) sentiment belief and its original (lexicon-based) polarity.  

In this learning manner, the proposed approach requires no prior knowledge or annotated data about a particular domain. It is thus applicable to multiple/different domains. 
Intuitively, this approach works based on two assumptions:

{\bf Assumption 1}: Sentiment Consistency~\cite{abelson1983whatever,liu2012sentiment}: a sentiment expression tends to be sentimentally coherent with its context. Notice that sentiment consistency can be reflected in multiple types of conjunction like ``and'', ``or'', etc., which will be explained in Section~\ref{sec:correlation}. In fact, this assumption is common in sentiment analysis and has been used in many studies~\cite{Kana:06,Ha:10} 

{\bf Assumption 2}: The number of domain polarity-changed lexical words is much smaller than the number of those whose polarities do not change. This assumption ensures that we can rely on the general-purpose lexicon itself for detection. 
In other words, the real polarity of a sentiment word in a certain domain can be distilled by its connections with other (mostly polarity-unchanged) words whose polarities are known from the lexicon.  

\subsection{Sentiment Word Collection}
As the first step, DSG collects sentiment words in an application domain corpus, including the sentiment words not in a sentiment lexicon. Specifically, we consider three types of (likely) sentiment words: (1) The word appears in a given lexicon. (2) The word is an adjective in the corpus. (3) The word is an adverb in the corpus and has an adjective form.

We simply accept all lexical words and adjective words as (likely) sentiment words, which does not cause serious problems in our experiments and they were also commonly used in the literature~\cite{liu2012sentiment}. However, we impose constraints on selecting adverbs. While adverbs like ``quickly'' and ``nicely'' do express sentiment, some others like ``very'' and ``often'' may not function the same. We thus use the adverbs having adjective forms only.

Notice that in the above setting, the sentiment words not in the lexicon are also collected due to two reasons: first, they are useful for building connection 
among other lexical words for inference purposes. Suppose that ``quiet'' is a sentiment word (found because it is an adjective) and it is not in the given lexicon. Given its sentiment correlations with other words like ``efficient and quiet'' and ``quiet and quick'', it can make a path between ``efficient'' and ``quick'' in the graph. Second, in each domain there exist a number of sentiment words uncovered by the given lexicon. Their inferred polarities can also benefit the graph reasoning process, though those words are not the focus in this study (we aim at detecting the polarity change of lexical words). For instance, if the non-lexical word ``quiet'' is identified as showing positive sentiment, ``efficient'' and ``quiet'' are more likely to be positive, given their in-domain sentiment correlations. 
We follow~\cite{das2007yahoo,Pa:08} to handle the negation problem, where a negation word phrase like ``not bad'' 
will be treated as a single word like ``not\_bad'' and its sentiment polarity will be reversed accordingly. Finally, all extracted words are modeled as nodes in the graph.

\subsection{Sentiment Correlation Extraction} 
\label{sec:correlation}
This step is to extract multiple types of conjunction relationship among sentiment words, which we refer to as \textit{sentiment correlation}\footnote{The term sentiment correlation used in this paper denotes the correlation between two sentiment words in a domain, which may not have to be the same as used in other studies.}. The key idea here is to use the sentiment consistency~\cite{abelson1983whatever,liu2012sentiment} (also see Assumption 1) for relationship construction among the collected sentiment words from the above step. Specifically, in an application domain, five (5) types of sentiment correlation are considered, each of which is presented in a triple format, denoted as ($word_1$, correlation type, $word_2$). They will be used in the subsequent graph inference (discussed in the next sub-section). Their definitions are shown in Table~\ref{tab:corraltion}. 

\begin{table*}[tb]
	\caption{\label{tab:corraltion} Five types of sentiment correlation.}
		\vspace{-2mm}
	\small
	\begin{center}
		\begin{tabular}{ccccc}
			\hline
			\textbf{Name} & \textbf{Correlation} & \bf Example  & \textbf{Representation} &  \bf Agreement Level\\
			\hline
			\hline
			\textbf{AND} & connecting with ``and'' & ``it is efficient and quiet" & (efficient, AND, quiet) & Strongly Agree\\
			\textbf{OR} & connecting with ``or'' & ``everything as expected or better" & (expected, OR, better) & Agree\\
			\textbf{NB} & neighboring words & ``a reasonably quiet fridge"  & (reasonably,  NB, quiet) & Weakly Agree\\
			\textbf{ALT} & although, though & ``too noisy, though it is efficient"  & (noisy, ALT, efficient) &  Disagree\\
			\textbf{BUT} & but, however  & ``it is a powerful but noisy machine"  & (powerful, BUT, noisy ) & Strongly Disagree\\
			\hline
		\end{tabular}
	\end{center}
	\vspace{-6mm} 
\end{table*} 

In each sentence
, when a specific type of relationship between two (collected) sentiment words is found, a triple is created. For instance, in the sentence ``it is efficient and quiet'', a triple (efficient, AND, quiet) will be generated. The extraction of $OR$ sentiment correlation is similar to $AND$. Likewise, a specific $BUT$ triple (powerful, BUT, noisy) will be extracted from the sentence ``it is a powerful but noisy machine''. The extraction of $ALT$ (abbreviation for although) is similar to $BUT$. $NB$ means two neighboring sentiment words in a sentence, like ``reasonably good''. 

Notice that while five types of relationships are jointly considered, they are associated with different agreement levels (parameterized in the graphical model discussed below). Here the agreement level measures how likely the sentiment polarities of two connecting words are the same. Intuitively, we believe $AND$ gives the highest level agreement. For instance, ``bad and harmful'' is very common but ``good and harmful'' is much unlikely. It is also an intuitive belief that $BUT$ indicates the strongest disagreement between two sentiment words. 
Note that we only consider pairwise relationships between sentiment words in this study, 
which already generate reasonably good results, as we will see.

\subsection{Graph Construction and Inference}
This subsection illustrates how our proposed domain-specific sentiment graph is constructed and used for word detection after the above two steps.

\subsubsection{Constructing Markov Random Field}
Markov Random Fields (MRFs) are a class of probabilistic graphical models that can deal with the inference problems with uncertainty in observed data. An MRF works on an undirected graph $G$, which is constructed by a set of vertexes/nodes $V$ and edges/links $E$ and denoted as $G = ( V, E )$. In the graph $G$, each node $v_i \in V$ denotes a random variable and each edge $(v_i, v_j) \in E$ denotes a statistical dependency between node $v_i$ and node $v_j$. Formally, $\psi_i(v_i)$ and $\psi_{ij}(v_i,v_j)$ are defined as two types of potential functions for encoding the observation (or \textit{prior}) of a node and the dependency between two nodes. They are also called \textit{node potential} and \textit{edge potential} respectively. An MRF thus can model a joint distribution for a set of random variables and its aim is to infer the marginal distribution for all $v_i \in V$. With an inference method used, 
the estimation of the marginal distribution of a node can be obtained, which is also called \textit{belief}.

The reason why we formulate our domain-specific sentiment graph as an MRF is three-fold: (1) The sentiment correlation between two words is a mutual relationship, as one word $w_a$ can provide useful sentiment information of the other word $w_b$ and vice versa, which can be properly formulated in an undirected graph. (2) From a probabilistic perspective, the polarity changes of sentiment words can be naturally understood as the belief estimation problem. That is, on one hand, we have an initial belief about the polarity of a lexical word (known from the lexicon, like the word ``cold'' is generally negative), which is essentially the prior. On the other hand, our goal is to infer the real polarity of a word in a specific application domain, which is reflected in its final estimated belief (like ``cold'' is positive in the domain \textit{fridge}). 
To be concrete, the polarity of a sentiment word is modeled as a 2-dimensional vector, standing for the probability distribution of positive and negative polarities, e.g., $\lbrack 0.9, 0.1 \rbrack$ indicates that a word is very likely to express a positive sentiment in an application domain. We can further use $p$ as the parameter to simplify the representation as $\lbrack p, 1-p \rbrack$. (3) Recall that multiple types of sentiment correlation are used and treated differently in our proposed approach, these \textit{typed sentiment correlations} can be encoded in the MRF model, which is further illustrated as follows.

\subsubsection{Inference over Typed Correlation}
As discussed above, the inference task in MRF is to compute the marginal distribution (or posterior probability) of each node given the node prior and edge potentials. Efficient algorithms for exact inference like Belief Propagation~\cite{pearl1982reverend} are available for certain graph topologies
, but for general graphs involving cycles the exact inference is computationally intractable. Approximate inference is thus needed. Loopy Belief Propagation~\cite{murphy1999loopy} is such an approximate solution using iterative message passing. A message from node $i$ to node $j$ is based on all message from other nodes to node $i$ except node $j$ itself. It works as:
\begin{equation} 
\small
\label{eq:message_passing0}
{m_{i \to j}}({x_j})=z\sum\limits_{{x_i} \in S} {{\psi _{i,j}}({x_i},{x_j})} {\psi _i}({x_i})\prod\limits_{k \in N(i)\backslash j} {{m_{k \to i}}({x_i})},
\end{equation}
where $S$ denotes the possible states of a node, i.e., being a sentiment word with positive or negative polarity. $x_j$ indicates that node $j$ is in a certain state. $N(i)$ denotes the neighborhood of $i$, i.e., the other nodes linking with node $i$. $m_{i \to j}$ is known as the message passing from node $i$ to node $j$. $z$ is the normalization constant that makes message ${m_{i \to j}}({x_j})$ proportional to the likelihood of the node $j$ being in state $x_j$, given the evidence from $i$ in its all possible states. 
After iterative message passing, the final belief $b_i(x_i)$ is estimated as: 
\begin{equation} 
\small
\label{eq:message_passing}
{b_i}({x_i}) = z'{\psi _i}({x_i})\prod\limits_{k \in N(i)} {{m_{k \to i}}({x_i})},
\end{equation}
where $z'$ is a normalization term that makes $\sum\nolimits_{{x_i}} {{b_i}({x_i}) = 1}$. In this case, ${b_i}({x_i})$ can be viewed as the posterior probability of a sentiment word being with positive or negative polarity.

However, notice that in the above setting, each edge is not distinguishable in terms of its type of sentiment correlation. In other words, each type of possible connections between words is treated intrinsically the same, which does not meet with our modeling needs. In order to encode the typed sentiment correlation as defined in previous sections, we propose to 
replace the Eq.~\ref{eq:message_passing} with:
\begin{equation} 
\vspace{-1mm}
\small
\label{eq:message_passing2}
\begin{split}
{m_{i \to j}}({x_j}) = & z\sum\limits_{{x_i} \in S} {\sum\limits_{{r_{i,j}} \in R} {{\psi _{i,j,{r_{i,j}}}}({x_i},{x_j})}}
\vspace{-2mm}\\
& {{\psi _i}({x_i})\prod\limits_{k \in N(i)\backslash j} {{m_{k \to i}}({x_i})} },
\end{split}
\end{equation}
where ${r_{i,j}} \in R$ indicates the specific type of sentiment correlation between node $i$ and node $j$, which can be any type like $AND$ or $NB$ as defined in Section~\ref{sec:correlation}. $\psi _{i,j,{r_{i,j}}}({x_i},{x_j})$ thus becomes an edge potential function related to its sentiment correlation type. Each type of a correlation is parameterized as a (state) transition matrix shown in Table~\ref{tab:transition}. The five types of sentiment correlation therefore result in five such tables but with different $\epsilon$ being set. For example, $\epsilon$ with $AND$ can be set to 0.3 as it indicates the highest agreement level, while the one with $NB$ can set to 0.1 as it is regarded as weakly agreement. For $BUT$, $\epsilon$ can be set to -0.3 as it shows strong disagreement.

\begin{table}[!htb]
			\vspace{-1mm}
	\caption{\label{tab:transition} Transition/Propagation matrix.}
		\vspace{-3mm}
	\small
	\begin{center}
		\begin{tabular}{ccc}
			\hline
			\textbf{State} & \textbf{Positive} & \textbf{Negative}\\
			\hline
			\hline
			\textbf{Positive} & $0.5 + \epsilon$ & $0.5 - \epsilon$ \\
			\textbf{Negative} & $0.5 - \epsilon$ & $0.5 + \epsilon$ \\
			\hline
		\end{tabular}
	\end{center}
	\vspace{-3mm} 
\end{table} 

For each word, with its estimated beliefs $\lbrack b^{+}, b^{-} \rbrack$ obtained in the application domain, its \textit{polarity change score (pcs)} is defined as:
 \vspace{-1mm}
\begin{equation}\label{eq:score}
pcs = I(l=+)b^{-} +I(l=-)b^{+},
\end{equation}
where $l$ denotes the original sentiment polarity of a lexical word, and $I(.)$ is the indicator function. According to the scores of all words, a word list ranked by $pcs$ is used to identify the most likely polarity-changed sentiment words, e.g., one can select the top $n$ words or set a threshold for word extraction. In this way, the sentiment words with changed polarities in each domain can be detected.

\section{Experiments}
\label{sec:experiments}
We conducted experiments on multiple datasets with several candidate solutions
. Here we first compare their performance on the word detection task. Sentiment classification will then be another task to evaluate their effect on polarity corrections.

\subsection{Experimental Setup}
\noindent
\textbf{Dataset.} Four (4) real-world datasets from different domains/products were used, namely, \textit{fridge}, \textit{blenders}, \textit{washer}, and \textit{movie}. The first three datasets contain review sentences provided by a collaborating business company. 
The fourth dataset (\textit{movie}) consists of tweets from Twitter discussing movies, which are collected by us. The first dataset contains 32,000 (32k) sentences, the second 16,000 (16k) sentences, and the rest datasets 10,000 (10k). Their data sizes can be viewed as large, medium and small. Such product diversity and variable size 
settings help evaluate the generality of each solution. Only the text is used by all candidate models. 

In addition, two other datasets from domains \textit{drill} and \textit{vacuum cleaner} are used as development sets for parameter selections. \textit{drill} contains 76k and \textit{vacuum cleaner} contains 10k sentences. 

\renewcommand{\tabcolsep}{1.5pt} 

\noindent
\textbf{Sentiment Lexicon.} We used a general-purpose sentiment lexicon from~\cite{Hu:04}, which contains 2,006 positive and 4,783 negative lexical words. A candidate model will find polarity-changed words from them for each domain.


\noindent
\textbf{Parameter Settings.} The hyper-parameters of state priors and the (typed) transition matrices in DSG are shown in Table~\ref{tab:prior} and~\ref{tab:relationship}. They were empirically set based on the word detection performance on the two development datasets. We found this parameter setting works generally well on both datasets, while they are from different domains and with different data sizes. 
The following reported results for evaluations are based on this setting and as we will see, it already produces quite good results. 

\renewcommand{\tabcolsep}{3pt} 


\begingroup
\begin{table}[!htb]
	\vspace{-1.5mm}
	\caption{\label{tab:prior} Parameters of state prior.}
		\vspace{-3mm}
	\small
	\begin{center}
		\begin{tabular}{cccc}
			\hline
			\textbf{Prior} & \textbf{Positive} & \textbf{Non-Lexical Words} & \textbf{Negative}\\
			\hline
			\hline
			\textbf{$p\; in\; \psi_i(v_i)$} & 0.70 & 0.50 & 0.30 \\
			\hline
		\end{tabular}
	\end{center}
	\vspace{-2mm} 
\end{table} 

\begin{table}[!htb]
	\vspace{-5mm}
	\caption{\label{tab:relationship} Parameters of typed transition matrix.}
		\vspace{-3mm}
	\small
	\begin{center}
		\begin{tabular}{c c c c c c}
			\hline
			\textbf{Types} & \textbf{AND}  & \textbf{OR} & \textbf{NB} & \textbf{ALT} & \textbf{BUT}\\
			\hline
			\hline
			\textbf{$\epsilon$ in $\psi_{i,j}(v_i, v_j)$} & 0.20 & 0.10 & 0.05 & -0.10 & -0.20 \\
			\hline
		\end{tabular}
	\end{center}
	\vspace{-5mm} 
\end{table} 
\endgroup

\subsection{Lexicon-based Sentiment Classifier}
Our evaluations include lexicon-based sentiment classification. We briefly illustrate how a lexicon-based sentiment classifier (called \textit{classifier} for short) 
works here. Clearly, it works with a lexicon, from which each word is associated with a sentiment score (e.g., -1/+1). The classifier then calculates the sentiment score $s$ of each sentence $t$ by summing the score of each word
. 
We follow the lexicon-based classifier design in~\citet{taboada2011lexicon}, incorporating sentiment negation and intensity. The sentence sentiment score $s$ is calculated: 
 \vspace{-5mm}
\begin{equation} 
 \vspace{-1.5mm}
\label{eq:output}
s = \sum\limits_{w \in t} {negation_w * intensity_w * polarity_w } \\
\end{equation}

Different lexicons working with this classifier will generate different results. That is, even if their lexical words are the same, the associated sentiment score of a lexical word 
could vary and Eq.~\ref{eq:output} will thus make different predictions. This is how we can utilize the classifier to verify the effect of the word detection
, because the classifier will perform differently using the original lexicon and the modified lexicon, whose results can be compared in a \textit{before-and-after} manner. Here the \textit{modified lexicon} means the sentiment polarities of (detected) lexical words are corrected from the original lexicon. For example, ``crush'' is associated with negative sentiment (-1) in the original lexicon, but it could be associated with positive sentiment (+1) in the modified lexicon (if detected), so the sentence-level sentiment scores will vary accordingly, e.g., ``the machine does crush ice!'' will be predicted as a positive sentence with the modified lexicon.

\subsection{Candidate Models}
\textbf{Original Lexicon (OL)}: This is a baseline for sentiment classification evaluations only (Section~\ref{sec:senti_cls}), which uses the classifier with the original lexicon. \\
\textbf{Domain-specific Sentiment Graph (DSG)}: This is our model. The following two models and it will be used for both word detection and classification.\\ 
\textbf{Lexicon-Classifier Inconsistency (LCI)}: This is a heuristic solution to detecting the polarity-changed sentiment words. It relies on the inconsistency between the sentiment of a lexical word (obtained from the original lexicon) and the sentiment of the sentences containing the word (
induced by the classifier). Concretely, it first calculates the sentiment polarities of all sentences using a classifier with the original lexicon. With the polarities of sentences known, it computes an inconsistency ratio for each lexical word. The inconsistency ratio is the ratio of (a) to (b), where (a) is the number of a word appearing in the positive/negative sentences but the word itself is negative/positive, and (b) is the number of all sentences covering that word. Finally, it ranks all lexical words based on their ratio values to produce a list of likely polarity-changed words. 
\\
\textbf{SentProp (SP)}: SentProp~\cite{hamilton2016inducing} is a lexicon construction algorithm considering the domain information, which is the most related work to ours. 
As discussed in Section~\ref{sec:related_work}, it is not directly applicable to the detection task. But since it can generate a list of domain-specific sentiment words and those words are associated with positive/negative scores (estimated by SentProp, which can be treated as the in-domain beliefs like DSG), we can design a two-step approach to achieve our goal.  First, we download\footnote{https://nlp.stanford.edu/projects/socialsent/} and run the SentProp system to learn the domain-specific lexicon for each domain. Second, we calculate the polarity change scores for all lexicon words like DSG based on the learned domain-specific sentiment scores and the original polarities from the lexicon using Eq.~\ref{eq:score}. Similar to DSG, it produces a 
list of words ranked by the polarity change scores. For its parameter selection, we tried both the system default, following the code instruction and the original paper, and also parameter fine-tuning based on the performance on two development sets (same as DSG), so as to achieve its best performance to report. 




\subsection{Correct Detection of Words}
\label{sed:correct_detection}
As each candidate model generates a list of words ranked by polarity-change scores, those top-ranked ones are the most likely polarity-changed words and can be used as the detected words. For evaluation, the top-$n$ words from each model are inspected and the number of correct (word) detection are counted, which is denoted as \textbf{\#C@n} in Table~\ref{tab:word_detection}.

Specifically, two domain experts who are familiar with the domain sentiments identify and annotate the correct polarity-changed words from the top-20 shortlisted candidate words generated by each model. For each candidate word, we sampled numbers of sentences containing that word for the domain experts to judge. A candidate word needs to be agreed by both of them to be correct. 
Here the Cohen's Kappa agreement score is 0.817. 

\renewcommand{\tabcolsep}{1.5pt} 

\begin{table}[!htb]
			\vspace{-1mm}
	\caption{\label{tab:word_detection} Detection of polarity-changed words.}
		\vspace{-2.5mm}
	\small
	\begin{center}
		\begin{tabular}{c l c c c c c}
			\hline
			\textbf{Model} & \textbf{\#C@n} & \textbf{fridge}  & \textbf{blender} & \textbf{washer} & \textbf{movie}\\
			\hline
			\hline
			& \#C@5 & 5  &  5  &  4 & 5\\
			DSG & \#C@10  & 9 & 10 &  6 & 9\\
			& \#C@20 & 12 &  15 &  12 & 15 \\
			\hline
			& \#C@5 & 3  & 3  & 3 & 1 \\
			LCI & \#C@10  & 5 & 3  & 5 & 4 \\
			& \#C@20 &  5 &   7   &  7 & 9\\
			\hline
			& \#C@5  & 1 & 0   & 1 & 1\\
			SP & \#C@10   & 2 & 0  & 2 & 4\\
			& \#C@20 &   3  &  3  &  3 & 6\\
			\hline
		\end{tabular}
	\end{center}
	\vspace{-5mm}
\end{table}

Evaluation results are reported in Table~\ref{tab:word_detection}, where we can see that DSG achieves outstanding results consistently. 
LCI also does a decent job, while SP does not perform well on this task. 

Next, we will evaluate the impact of such detection from their top 20 words, and the following sub-sections are based on their correctly detected words to give further analyses.


\subsection{Sentiment Classification}
\label{sec:senti_cls}

After the detection of polarity-changed words, we conduct classification tasks on the sentences containing at least one word from the detected words of all models. Because the classification results on the sentences that without containing any detected word would not be affected (same prediction results using either the original or modified lexicon). 

For evaluation, we sampled and labeled 925 (around 1k) sentences, from all sentences that could be affected. We used stratified sampling strategy, and meanwhile, set minimum number of sentences contained by each word, to make sure each detected word is considered. 
The numbers of labeled sentences for the four domains are 232, 214, 174, and 305. The Cohen's Kappa agreement score is 0.788.

In regard to the lexicon-based classification, for DSG and LCI, the modified lexicon for each domain is based on the correction of the original lexicon (OL) on that domain. For SP, its self-generated lexicon is used with its inferred sentiment scores.

\begin{table}[!htb]
	\vspace{-1mm}
	\caption{Sentiment classification accuracy.}\label{tab:classification}
	\small
		\vspace{-2mm}
	\begin{center}
		\begin{tabular}{c c c c c c c}
			\hline
			\textbf{Model} & \textbf{fridge}  & \textbf{blender}    & \textbf{washer} & \textbf{movie} & \textbf{AVG}\\
			\hline
			\hline
			DSG &	\textbf{74.56\%}	& \textbf{80.84\%}		&	\textbf{77.01\%}	&	84.91\%	&	\textbf{79.33\%} \\			
			SP &	 68.10\%	&	78.97\% 	&	66.67\%	&	\textbf{87.87}\%	&	75.40 \%	\\
			LCI &	61.63\%	&	68.22\%		&	62.64\%	&	62.95\%	&	63.86\%	\\
			\hline
			OL  &	61.20\%  & 65.42\%	&	62.06\%	&	56.72\%	&	61.35\%	\\
			\hline
		\end{tabular}
	\end{center}
		\vspace{-3mm}
\end{table}

Table~\ref{tab:classification} reports the classification accuracy, from which we have the following observations: 
\begin{enumerate}
	\item  Compared to the baseline using the original lexicon (OL), DSG greatly improves the accuracy by 17.98\% in average. 
	We can see the usefulness of detecting polarity change of lexical words for sentiment classification.   
	
	\item  SP also produces very good results. 
	The reason is, as an essentially lexicon-generation approach, SP itself creates a bigger lexicon for each domain (around 2 times bigger than OL), including additional sentiment words outside the original lexicon. In other words, discovering more sentiment words (with more sentiment clues provided) could also help better classification. Note that this does not contradict the importance of detecting polarity-change words, as they are two different aspects. We will further discuss this shortly.
	
	\item LCI outperforms OL 
	but its performance gain is low. The reason is, though LCI detects polarity changed words decently, its detected words affect a much smaller number of sentences compared to DSG's and SP's, i.e., the words LCI detects are rarer and less frequent, with fewer sentences being affected.
	

\end{enumerate}

\subsection{Example of Polarity Change in Domains}
Here we show some example results. An underlined word indicates its polarity has changed in a certain domain. We found ``good for \underline{crushing} ice'' and ``this \underline{breaks} down frozen fruit'' in \textit{blender}, ``\underline{damn}, I wanna watch it so \underline{bad}!'' and ``this movie was \underline{insanely} brilliant!'' in \textit{movie}. We also found 
``it keeps things \underline{cold} and \underline{frozen}'' in \textit{fridge} and ``you can also \underline{delay} your cycle'' in \textit{washer}. 


\subsection{Further Analysis and Discussion}
We aim to answers two questions here. Q1: What is the key difference between using SP and DSG? Q2: More generally, what is the relationship between the existing lexicon generation research and this polarity-change detection problem?


First, let us deep dive a bit into SP. As a lexicon generation approach, its goal is to collect sentiment words from a given corpus and infer their sentiment scores. There are two important notes: (a) while SP could discover more sentiment words, those extracted words could be wrong. For example, SP extracts the word ``product'' as a sentiment word and assigns it a positive (+ve) sentiment. This could lead to mis-classifications of the negative (-ve) sentences containing ``product''. (b) while SP directly discovers and estimates sentiment words, it does not know which sentiment words carry domain-oriented important information.
For example, SP discovers ``excellent'', ``crush'', ``terrible'' for the domain \textit{blender} and estimates the sentiment scores as 0.9, 0.7, and 0.1 (for simplicity, let us assume all scores are rescaled to [0.0, 1.0] , where 1.0 denotes most +ve and 0.0 most -ve). Those scores indicate their polarities, but do not reflect their importance/effect of polarity change towards a domain.


For DSG, (a) could be avoided because ``product'' is usually excluded in a general-purpose lexicon. Regarding (b), say the scores of ``excellent'', ``crush'' and ``bad'' are 1.0, 0.0, and 0.0 in the original lexicon, with the domain sentiment re-estimation from DSG they become 0.9, 0.7, and 0.1. Their polarity-changed scores are thus inferred as 0.1 ($|1.0-0.9|$), 0.7, and 0.1, where ``crush'' as an important domain-sentiment changed word (0.7) can be found.

Certainly, one can compare the SP generated lexicon to the original/general lexicon. We already did this for the detection task (
Table~\ref{tab:word_detection}). Here we design a variant {SP-dsg-like}, following this idea for the classification task. The main difference between SP and SP-dsg-like is that SP directly uses its generated lexicon and sentiment scores, while SP-dsg-like uses its generated lexicon to modify the original lexicon (OL) like DSG does.  However, SP-dsg-like performs poorly (Table~\ref{tab:classification2}), mainly because the modified lexicon (based on OL) does not fully reflect the whole sentiment words generated by SP.

\begin{table}[!htb]
	\vspace{-1mm}
	\caption{Sentiment classification accuracy.}\label{tab:classification2}
	\small
	\vspace{-2mm}
	\begin{center}
		\begin{tabular}{c c c c c c c}
			\hline
			\textbf{Model} & \textbf{fridge}  & \textbf{blender}  & \textbf{washer} & \textbf{movie} & \textbf{AVG}\\
			\hline
			\hline
			OL  &	61.20\%  & 65.42\%	&	62.06\%	&	56.72\%	&	61.35\%	\\
			\hline
			SP &	 68.10\%	&	78.97\% 	&	66.67\%	&	87.87\%	&	75.40\%	\\
			\hline
			
			SP-dsg-like &	67.67\% &	71.02\%	& 64.36\%	&	63.93\%	&	66.75\% \\ 
			SP-dsg-like +SP &	69.40\% &	77.57\%	&  68.97\%	&	83.28\%	&	74.81\% \\ 
			OL + SP  &	 62.07\%	&	78.04\%	 &	70.11\%	&	82.30\%	&	73.13\%	\\
			CLI + SP   &	 62.07\%	&	77.57\%	 &	70.67\%	&	83.61\%	&	73.48\%	\\
			DSG + SP  &	 \textbf{72.41\%}	&	\textbf{78.97\%}	 &	\textbf{79.89\%}	&	\textbf{88.85\%} & \textbf{80.03\%}	\\
			\hline
		\end{tabular}
	\end{center}
	\vspace{-4mm}
\end{table}

We then combine two lexicons together to give another variant SP-dsg-like+PS, which means the modified lexicon (based on OL) is expanded by the SP self-generated lexicon, where the SP generated lexicon can contain additional sentiment words (outside OL). Similarly, we can make OL+SP and CLI+SP, but they are all inferior to SP (Table~\ref{tab:classification2}). The reason is that the key polarity-changed words in the original lexicon have not been corrected, keeping wrong sentiments for classification tasks.


However, noticing DSG can effectively detect and correct those words, when we use DSG+SP, the overall results are improved (AVG) and even better than using DSG or SP only (Table~\ref{tab:classification2} and Table~\ref{tab:classification}). 
 
 


%
%



It has been demonstrated that either more sentiment words (from PS) or fixing a small number of important polarity-changed words (from DSG) can help sentiment classification. With DSG+PS working better, we can view the lexicon generation and domain polarity word detection as two directions for classification improvement. Either one has its own advantage. The lexicon generation approaches can induce more words and may help find rarer/infrequent words. The word detection can be handy and less risky, as it simply corrects the polarities of some important lexical words 
and would not induce noises (wrong sentiment words).  

Finally, the answers to Q1 and Q2 are: using SP/lexicon-generation and DSG/polarity-change-detection can both improve sentiment classification, but in different manners. Using DSG can also effectively detect important polarity-change words, while SP does not perform very well on this task. These two directions could be complementary, as shown by DSG+SP. It is our hope that this work can inspire further relevant research in the future.


\section{Conclusion}
This paper studied the problem of detecting domain polarity-changed words in a sentiment lexicon. 
As we have seen, the wrong polarities seriously degenerate the sentiment classification performance. To address it, this paper proposed a novel solution named Domain-specific Sentiment Graph (DSG). 
Experimental results demonstrated its effectiveness in finding the polarity-changed words and its resulting performance gain in sentiment classification.
%
%

\bibliography{domain_sentiment_short}
\bibliographystyle{acl_natbib}

\end{document}